%% file: main.tex
\documentclass[letterpaper, 10 pt, conference]{ieeeconf}

\IEEEoverridecommandlockouts                              
\usepackage{amsmath,amsfonts}
\usepackage{algorithmic}
\usepackage{algorithm}
\usepackage{array}
\usepackage[caption=false,font=normalsize,labelfont=sf,textfont=sf]{subfig}
\usepackage{textcomp}
\usepackage{stfloats}
\usepackage{url}
\usepackage{verbatim}
\usepackage{graphicx}
\usepackage{cite}
\usepackage{booktabs}
\usepackage{multirow} 
\usepackage[table]{xcolor}
\usepackage{wrapfig}
\usepackage{makecell}

\usepackage{xcolor}

\begin{document}

\title{\LARGE \bf
       \textbf{DUET-DINO}: Simultaneous Cross-View World Modeling for Latent Planning in Robot Manipulation
       
        
    }

\author{
        Nisarga Nilavadi$^{1}$,
        Ralf Römer$^{2}$,
        Moritz Reuss$^{3,4}$,
        Michael Krawez$^{1}$, 
        Tobias Jülg$^{1}$, \\
        Angela P. Schoellig$^{2,5}$,
        Rudolf Lioutikov$^{3,5}$,
        Wolfram Burgard$^{1,5}$
\thanks{
$^{1}$Artificial Intelligence and Robotics Lab, University of Technology Nuremberg (UTN), Germany.}
\thanks{$^{2}$Learning Systems and Robotics Lab, Technical University of Munich (TUM), Germany.}
\thanks{$^{3}$Intuitive Robots Lab, Karlsruhe Institute of Technology (KIT), Germany.}
\thanks{$^{4}$NVIDIA.}
\thanks{$^{5}$Robotics Institute Germany.}
}



\newcommand{\algname}{DUET-DINO}
\newcommand{\std}[1]{{\scriptsize$\pm$$#1$}}

\maketitle

\input{1_abstract}

\input{2_introduction}

\input{3_relatedwork}

\input{4_method}

\input{5_experiments}

\input{6_conclusion}

\bibliographystyle{IEEEtran}
\bibliography{sources.bib}

\end{document}

%% file: 1_abstract.tex
\begin{abstract}
Action-conditioned latent world models predict future visual representations, enabling zero-shot goal-conditioned robot planning and control.
However, their predictions for fine-grained spatial and rotational actions are unreliable for full 7-DoF end-effector control.
To address this gap, we introduce \algname{}, a simultaneous cross-view latent world model that jointly learns action-conditioned predictions from static side- and wrist-camera observations through cross-view conditioning. 
By exploiting complementary global scene and gripper-centric information, \algname{} enables latent planning over the full 7-DoF action space. 
Across spatially diverse reach, orientation-intensive angled-reach, and multi-goal grasp-and-lift tasks, \algname{} consistently outperforms single-view and independent dual-view baselines, achieving $\mathbf{92\%}$ success on reach, $\mathbf{72.5\%}$ on angled-reach, and $\mathbf{60.0\%}$ on lift tasks.
DUET-DINO is trained from scratch on DROID and RoboArena datasets and generalizes robustly under visual distribution shifts.
We further show that while V-JEPA~2 wrist-view predictions underestimate visual dynamics induced by fine-grained actions, DINOv3 predictions better capture action-conditioned scene changes, leading to stronger downstream planning.
The code and model checkpoints will be open-sourced.
Project page: \textcolor{teal}{\url{https://utn-air.github.io/DUET-DINO}}
\end{abstract}

%% file: 2_introduction.tex
\section{Introduction}
\label{sec:intro}

World models have emerged as a promising direction to address generalization in robotics by learning representations of the world and its dynamics from large-scale web data~\cite{agarwal2025cosmos, assran2025vjepa2, hou2026world}.
Through self-supervised training on billions of examples, world models can improve generalization across diverse environments and implicitly capture aspects of real-world dynamics, such as how objects move and interact.
In robot manipulation, world models can predict future states as latent representations~\cite{assran2025vjepa2, murlabadia2026vjepa21}, pixel-level observations~\cite{agarwal2025cosmos, guo2025ctrl, kim2026cosmos}, or rewards~\cite{kim2026cosmos}, thereby providing information about the consequences of actions. 
These predictions can be leveraged in various ways, such as for planning~\cite{assran2025vjepa2, terver2025drives}, policy training~\cite{chandra2025diwa, jiang2025world4rl}, or safety verification~\cite{li2025worldeval, team2025evaluating}.
Latent world models~\cite{assran2025vjepa2, murlabadia2026vjepa21, zhou2025dino} are particularly attractive for closed-loop planning and control: they produce task-relevant features without modeling photorealistic details, enabling much faster rollouts than pixel-space generators.

\begin{figure}[t]
\centering
\includegraphics[width=1\linewidth]{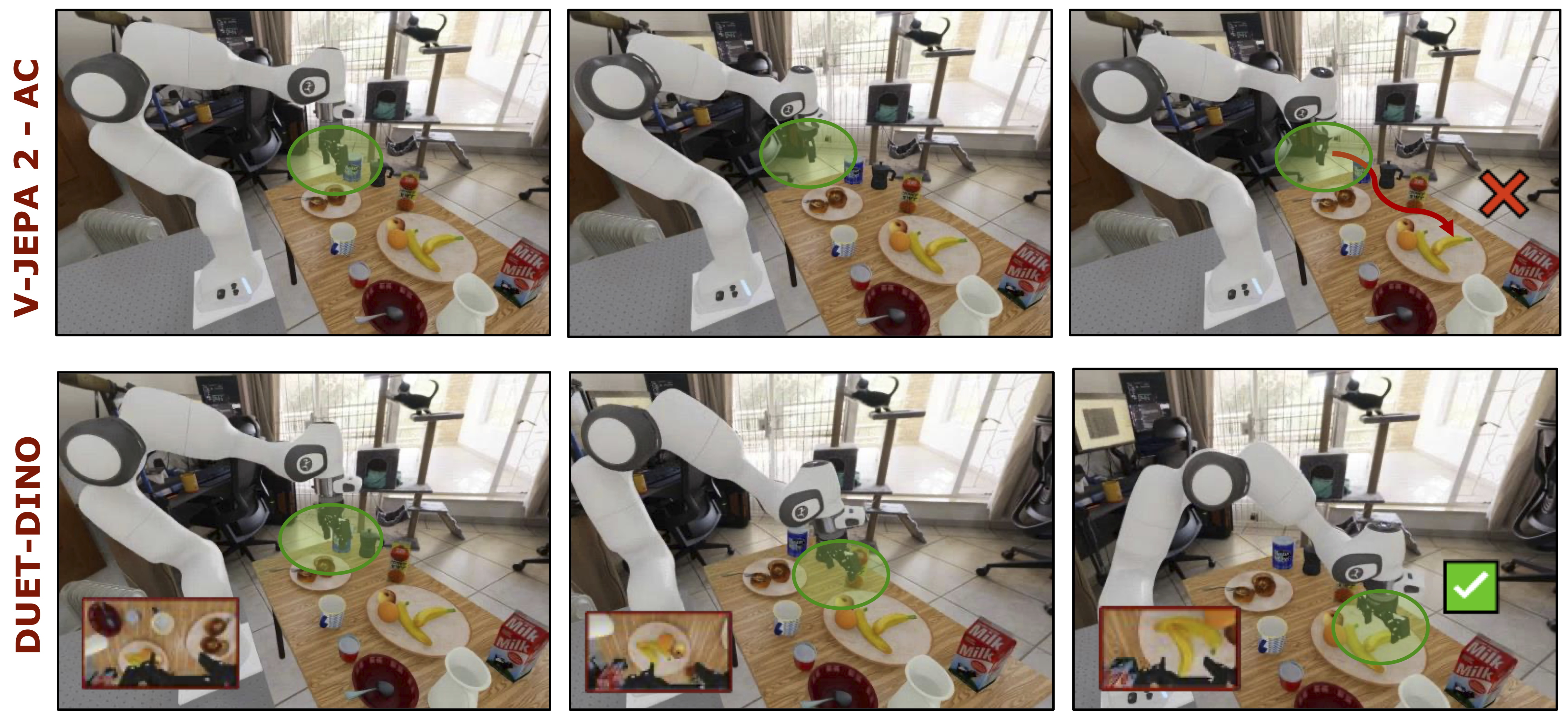}
\caption{Comparison of goal-conditioned 7-DoF action planning toward the target object (banana) between our proposed DUET-DINO world model and V-JEPA~2-AC. DUET-DINO progressively moves the end-effector toward the goal, while V-JEPA~2-AC fails to approach the target.}
\label{fig:cover_fig}
\end{figure}

Despite these compelling properties, applying latent world models to real-world manipulation remains challenging. 
While existing action-conditioned predictors~\cite{assran2025vjepa2,murlabadia2026vjepa21} can be conditioned on the full 7-DoF end-effector action space, their rollouts primarily capture coarse translational motion over short horizons. 
Accordingly, prior deployment setups have focused on translation-only action planning from predicted rollouts for relatively simple pick-and-place downstream tasks. 
In contrast, full 7-DoF action planning requires reliable prediction of fine-grained translational, rotational, and gripper-state changes, which are critical for precise manipulation tasks such as spatially diverse reaching, angled grasping, orientation-intensive pick-and-place, and insertion~\cite{terver2025drives}. 
Consequently, tasks that require planning over the full end-effector action space remain out of reach for current static single-view latent world models.

To address this gap, we introduce a tightly modeled dual-view approach, \textbf{Simultaneous Cross-View World Modeling for Latent Planning in Robot Manipulation (\algname{})}, for planning full 7-DoF end-effector motions.
\algname{} goes beyond prior latent world models~\cite{assran2025vjepa2,murlabadia2026vjepa21} by jointly modeling scene changes from a side-view camera that captures global workspace context and a wrist-mounted camera that observes gripper-centric geometry.
The two streams are conditioned on each other in the latent space via cross-attention, enabling each view predictor to exploit complementary cues from the other while remaining specialized to its own camera.
At deployment, \algname{} plans over full 7-DoF action sequences by jointly minimizing the dual-view goal cost across both latent streams.
On hardware and simulated benchmarks including spatially diverse, orientation-intensive, and multi-goal angled-lift tasks, \algname{} consistently outperforms static single-view and naive dual-view baselines.

In summary, our main contributions are the following:
\begin{enumerate}
    \item We propose \algname{}, a latent world model that jointly trains side- and wrist-view action-conditioned predictors on cross-view-conditioned observation latents, enabling each view-specific predictor to leverage complementary information for improved downstream planning.
    \item We extend latent world-model planning to the full 7-DoF end-effector space by jointly optimizing over predicted side and wrist-view latent rollouts using a normalized dual-view goal cost.
    \item We show the effect of pretrained latent representations on action-conditioned predictions, finding that V-JEPA~2 underestimates wrist-view visual changes, while DINOv3 better captures fine-grained motion.
    \item We introduce spatially diverse, orientation-intensive, and multi-goal manipulation tasks in the RoboLab simulator~\cite{yang2026robolab} that expose the 7-DoF prediction bottlenecks of latent world models.
    \item We conduct extensive evaluations across simulation and hardware on spatial reach, orientation-intensive angled reach, and sequential angled-grasp and lift tasks, including diverse backgrounds, distractors, and perturbations. 
    \algname{} achieves $92\%$ success on reach, $72.5\%$ on angled-reach, and $60\%$ on lift tasks, consistently outperforming single-view and independent dual-view baselines.
\end{enumerate}

%% file: 3_relatedwork.tex
\section{Related Work}
\label{sec:citations}

\subsection{World Models in Robotics}
\label{sec:wm_literature}


\textbf{Generative video world models.}
Foundation-scale video generators such as Cosmos~\cite{agarwal2025cosmos} and Wan~\cite{wan2025wan} use diffusion- and flow-based models to synthesize photorealistic video, and a growing number of works adapt this capability to robot manipulation.
DreamGen~\cite{jang2025dreamgen} synthesizes neural trajectories for policy training, recovering latent pseudo-actions to facilitate generalization.
WorldGym~\cite{quevedo2025world} uses an autoregressive video generator as a proxy environment for scoring policies via Monte Carlo rollouts, while Ctrl-World~\cite{guo2025ctrl} combines pose-conditioned memory with multi-view video diffusion to rank policies and produce imagined rollouts for supervised fine-tuning.
DiWA~\cite{chandra2025diwa} pushes this further to fully offline reinforcement learning for adapting diffusion policies inside a learned world model.
While generative world models excel at visual fidelity, reconstructing every pixel is computationally expensive and prone to hallucinations of physically implausible scenes~\cite{mei2025world}.

\textbf{Latent world models.}
A complementary line of work avoids pixel reconstruction by predicting representations in a learned latent space.
Joint-Embedding Predictive Architectures (JEPAs), as introduced for images by I-JEPA~\cite{assran2023self} and extended to video by V-JEPA~\cite{bardes2024revisiting}, leverage masked feature-prediction objectives to produce strong, transferable representations without any reconstruction or hand-crafted augmentation.
V-JEPA 2~\cite{assran2025vjepa2} scales this recipe further and adds an action-conditioned predictor (V-JEPA 2-AC) using under 62 hours of robot interaction.
V-JEPA 2.1~\cite{murlabadia2026vjepa21} introduces a dense predictive loss and deep self-supervision to better capture fine-grained features.
DINO-WM~\cite{zhou2025dino} demonstrates that patch features from a frozen DINOv2 encoder can support action-conditioned latent prediction and zero-shot planning, and DINOv3~\cite{simeoni2025dinov3} pushes dense self-supervised features to a larger scale.
A key limitation of existing latent world models is that they largely rely on a single external camera and struggle with fine-grained 7-DoF control, an open challenge that we explicitly target in this work.

\subsection{Visual Planning with World Models}
\label{sec:wm_visualplanning}


\textbf{Planning in pixel vs. latent space.}
One family of methods plans directly in pixel space by generating future videos and extracting actions from them.
UniPi~\cite{du2023learning} casts decision making as text-conditioned video generation followed by an inverse-dynamics model; SuSIE~\cite{black2024zero} uses a pretrained image-editing diffusion model to propose subgoal images for a low-level goal-conditioned policy; and CLOVER~\cite{bu2024closed} closes the loop by combining a text-conditioned video diffusion model with a measurable error space and a feedback-driven controller.
These methods directly couple planning quality to the fidelity of generated pixels.
An alternative is to plan in the latent space of a latent world model: V-JEPA 2-AC~\cite{assran2025vjepa2} and DINO-WM~\cite{zhou2025dino} formulate manipulation as latent-feature matching to a goal embedding and optimize action sequences directly against this latent cost.
Wang et al.~\cite{wang2026temporal} show that explicitly regularizing latent trajectories to be locally straight improves the geometry of the latent space and makes Euclidean distances a better proxy for cost.

\textbf{Action proposals and optimization.}
A common choice is to sample actions from a Gaussian proposal and iteratively refine them using the cross-entropy method (CEM)~\cite{rubinstein1997optimization, bharadhwaj2020model, nagabandi2020deep, zhou2025dino, assran2025vjepa2}, which is derivative-free and well-suited to highly non-convex objectives.
Other works learn structured proposal distributions~\cite{gao2025flip} or rely on a vision-language model to propose camera trajectories for spatial reasoning~\cite{yang2025mindjourney}.
Optimization can also blend sampling and gradients: \cite{bharadhwaj2020model} interleave CEM with gradient steps to escape local optima, and \cite{nagabandi2020deep} shows that simple MPPI-style updates already enable dexterous in-hand manipulation when paired with sufficiently accurate dynamics.
Recent works plan from a single external camera and primarily address translation-dominated tasks.
Rotational and gripper actions remain challenging because their effects are degraded in latent predictions~\cite{assran2025vjepa2,terver2025drives}.
In addition, goal cost is poorly informed by a single global view.
\algname{} addresses this bottleneck by conditioning the global side- and gripper-centric wrist observation views on each other in latent space, and performing CEM planning with a goal cost that aggregates both views, enabling reliable planning over the full 7-DoF action space.

%% file: 4_method.tex
\section{Methodology}
\label{sec:approach}

\begin{figure*}[t]
\centering
\includegraphics[width=0.9\linewidth]{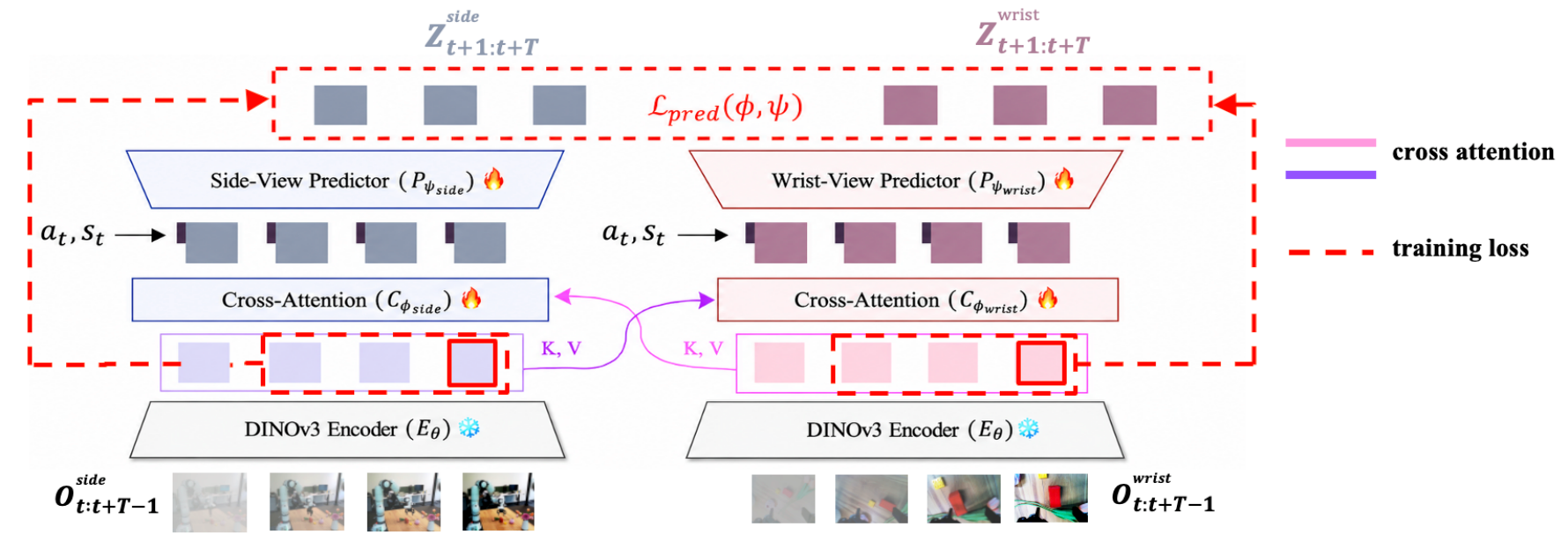}
\caption{
Architecture and training of \algname{}.
The pretrained encoder maps side- and wrist-view observations into latent space.
Cross-attention blocks condition each view on complementary information from the other view in the latent space.
The view-specific predictor heads then predict future latents conditioned on the action and end-effector state.
The dual-view prediction loss $\mathcal{L}_{\mathrm{pred}}(\phi,\psi)$ combines the teacher-forcing loss computed over all one-step predictions in the clip with an autoregressive loss over the $K$-step rollout predictions.
}
\label{fig:network}
\end{figure*}


\subsection{Problem Formulation}
\label{sec:problem}
We approach tabletop manipulation using synchronized camera streams from a static side camera and a dynamic wrist-mounted camera.
Given an observation~${o_t=(o_t^{\mathrm{side}}, o_t^{\mathrm{wrist}})}$ consisting of two camera images~${o_t^{\mathrm{side}}, o_t^{\mathrm{wrist}} \in \mathbb{R}^{H \times W \times C}}$, as well as the full end-effector action ${a_t \in \mathbb{R}^{7}}$ at timestep~$t$, we aim to predict future latent representations~$\hat{z}_{t+T}^{\mathrm{side}},\hat{z}_{t+T}^{\mathrm{wrist}}$ induced by the executed actions over horizon~$T$.
The predicted latents should capture visual features from both camera views, enabling zero-shot goal-conditioned planning for manipulation.

\subsection{\algname{}: Simultaneous Cross-View World Modeling}
\label{sec:dw-worldmodel}

The proposed world model consists of three components.
First, a pretrained visual encoder maps side- and wrist-view images to latent representations.
Next, cross-attention blocks allow each view-specific latent to query the complementary-view latents.
Finally, using the updated latent representations, separate predictor heads predict future latents for their respective views, conditioned on the same robot action.
We keep the visual encoder frozen and train the cross-attention blocks and predictor heads jointly from scratch on large-scale robotic manipulation datasets. 
An overview of \algname{} is shown in Fig.~\ref{fig:network}.

The side- and wrist-view observations $o_t^{\mathrm{side}}$ and $o_t^{\mathrm{wrist}}$ are independently encoded using the frozen DINOv3 encoder ${E}_\theta$ which was pretrained on large-scale visual data, yielding patch-wise latent representations
\begin{equation}
    z_t^{v} = E_\theta(o_t^{v}) \in \mathbb{R}^{P \times D}
    \qquad v \in \{\mathrm{side}, \mathrm{wrist}\}.
\end{equation}
where $P$ is the number of image patches, and $D$ is the encoder's latent dimension per patch.
We then feed these latents through learnable cross-attention blocks, $\mathcal{C}_{\phi_{\mathrm{side}}}$ and $\mathcal{C}_{\phi_{\mathrm{wrist}}}$, where the latents of each view query the latents of the other view.
Within each cross-attention block, the target-view latents serve as queries, while the other-view latents serve as keys and values
\begin{equation}
\begin{aligned}
    \tilde{z}_t^{\mathrm{side}}
    &=
    \mathcal{C}_{\phi_{\mathrm{side}}}
    \left(
        Q=z_t^{\mathrm{side}},
        K=z_t^{\mathrm{wrist}},
        V=z_t^{\mathrm{wrist}}
    \right),
    \\
    \tilde{z}_t^{\mathrm{wrist}}
    &=
    \mathcal{C}_{\phi_{\mathrm{wrist}}}
    \left(
        Q=z_t^{\mathrm{wrist}}, 
        K=z_t^{\mathrm{side}},
        V=z_t^{\mathrm{side}}
    \right).
\end{aligned}
\end{equation}
The resulting cross-view-conditioned latents, $\tilde{z}_t^{\mathrm{side}}$ and $\tilde{z}_t^{\mathrm{wrist}}$, are passed to their respective state- and action-conditioned predictor heads, $P_{\psi_{\mathrm{side}}}$ and $P_{\psi_{\mathrm{wrist}}}$, for which we adopt the architecture introduced by~\cite{assran2025vjepa2}.
The predictor heads remain view-specific, with their inputs having already been updated through cross-view latent conditioning.
This allows each predictor to preserve its own view-specific dynamics while using complementary information from the other view to predict the next latent representation as
\begin{equation}
    \hat{z}_{t+1}^{v}
    =
    P_{\psi_v}
    \left(
        \tilde{z}_t^{v}, a_t, s_t
    \right)
    \qquad
    v \in \{\mathrm{side}, \mathrm{wrist}\}.
\end{equation}

\textit{Training:}
We train the cross-attention blocks and predictor heads of \algname{} from scratch on a large-scale dataset~$\mathcal{D}=\{(o_{t:t+T}^i,a_{t:t+T-1}^i)\}_{i=1}^{N}$ containing clips of length~$T$ of observation-action pairs.
For both camera views~$v \in \{\mathrm{side}, \mathrm{wrist}\}$, we combine two loss functions for training: a teacher-forcing (one-step prediction) loss $\mathcal{L}_{\mathrm{TF}}^{v}$ and an autoregressive rollout loss $\mathcal{L}_{\mathrm{AR}}^{v}$.
We compute the latter over a rollout horizon of $K=2$, starting from the first frame of the clip~\cite{assran2025vjepa2}.
The overall dual-view prediction loss for joint training of the cross-attention blocks and the view-specific predictor heads is given by
\begin{equation}
\begin{gathered}
    \label{eq:training_loss}
    \mathcal{L}_{\mathrm{pred}}(\phi,\psi)
    =\mathbb{E}_{(o_{t:t+T},a_{t:t+T-1})\sim \mathcal{D}} \!\!\!\!\!\!
    \sum_{v \in \{\mathrm{side}, \mathrm{wrist}\}}
    \\
    \Bigg[
    \underbrace{
    \frac{1}{T-1}
    \sum_{t=2}^{T}
    \left\|
    \hat{z}_{t}^{v} - z_{t}^{v}
    \right\|_1
    }_{\mathcal{L}_{\mathrm{TF}}^{v}}
    +
    \underbrace{
    \frac{1}{K}
    \sum_{k=2}^{K+1}
    \left\|
    \hat{z}_{k,\mathrm{AR}}^{v} - z_{k}^{v}
    \right\|_1
    }_{\mathcal{L}_{\mathrm{AR}}^{v}}
    \Bigg].
\end{gathered}
\end{equation}

\subsection{Visual Planning}
\label{sec:worldmodel_planning}
Once trained, we leverage \algname{} for zero-shot planning of full end-effector motions.
Our objective is to generate a sequence of actions driving the robot to a goal observation~$o_{\text g}=(o_{\text g}^{\mathrm{side}},o_{\text g}^{\mathrm{wrist}})$.
We formulate this problem in the dual learned latent space of \algname{}, with the goal embeddings given by~$z_{\text g}^{\mathrm{side}}=E_\theta(o_{\text g}^{\mathrm{side}})$, $z_{\text g}^{\mathrm{wrist}}=E_\theta(o_{\text g}^{\mathrm{wrist}})$.
Denoting~$z_t=(z_t^{\mathrm{side}},z_t^{\mathrm{wrist}})$ and considering a fixed receding horizon of length~$H$, we aim to iteratively solve the optimal control problem
\begin{align}
\begin{split}
    \label{eq:ocp}
    \min_{a_{0:H-1}} \;\, J(z_{\text g},\,&\hat{z}_{H})
    \\ \text{s.t.} \quad\; \hat{z}_{\tau + 1} &= \text{DUET-DINO}(\hat{z}_\tau,a_\tau), \; \tau =0,\dots,H-1, \\
    \hat{z}_0 &= z_t,
\end{split}
\end{align}
where~$J(z_{\text g},\hat{z}_{H})$ measures the distance between the predicted and goal latents.
We solve the optimal control problem~\eqref{eq:ocp} using CEM~\cite{assran2025vjepa2, terver2025drives}.
At each CEM iteration, we sample~$N$ action sequences~$a_{t:t+H-1}^{1:N} \sim \mathcal{N}(\mu,\Sigma)$ from a Gaussian proposal distribution with mean~$\mu$ and variance~$\Sigma$ and roll out the actions inside \algname{} to obtain imagined latent trajectories~$\tau^{1:N} = (s_t,z_t,a_t,\hat{z}_{t+1}^{1:N},a_{t+1}^{1:N},\dots,\hat{z}_{t+H}^{1:N})$.
Then, we can assess the quality of each action sequence by computing the corresponding deviations from the goal latents as
\begin{equation}
\begin{gathered}
    \label{eq:goal_deviations}
    \ell^\mathrm{side}_i
    \!= \|\hat{z}_{t+H}^{\mathrm{side},i}-z_{\text g}^\mathrm{side}\|_1, \;\;
    \ell^\mathrm{wrist}_i
    \!= \|\hat{z}_{t+H}^{\mathrm{wrist},i}-z_{\text g}^\mathrm{wrist}\|_1.
\end{gathered}
\end{equation}
The overall dual-view planning cost is obtained by summing the two per-view costs and normalizing by the dual-view cost of a zero-action rollout over the same prediction horizon, i.e.,
\begin{equation}
    J_i(z_{\text g},\hat{z}_{t+H})
    =
    \frac{
        \ell_i^{\mathrm{side}}+\ell_i^{\mathrm{wrist}}
    }{
        (\ell^{\mathrm{side}}+\ell^{\mathrm{wrist}})_0
    }
    \qquad
    \mathcal{J}=\{J_i\}_{i=1}^{N}.
\label{eq:mpc_cost}
\end{equation}
We adopt a weighted top-$k$ CEM update, where the selected candidates are weighted using a softmax over their negative costs, assigning larger weights to lower-cost action sequences when updating the proposal distribution.


%% file: 5_experiments.tex
\section{Experiments}
\label{sec:results}

\begin{figure*}[t]
\centering
\includegraphics[width=\linewidth]{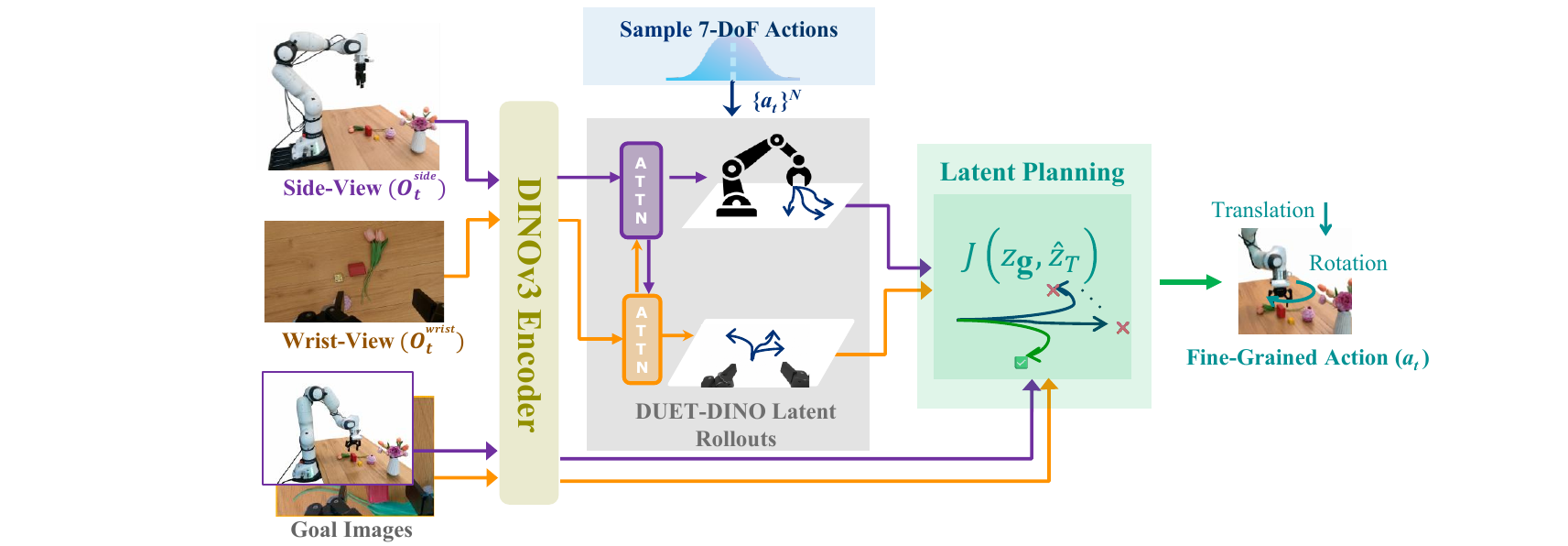}
\caption{
    Overview of the proposed simultaneous cross-view world modeling method, showing predicted dual-view latent rollouts, CEM action planning using predicted and goal latents, and the normalized dual-view planning cost $J(z_\mathrm{g},\hat{z}_T)$.
}
\label{fig:latent_planning}
\end{figure*}

This section describes the dataset processing, world-model training procedure, experimental setup, evaluation protocol, and results for \algname{} and the baseline methods.

\subsection{Implementation and Training Details}
\label{sec:datasets}
We train and evaluate our world-model predictors on two large-scale robot manipulation datasets: DROID~\cite{khazatsky2024droid} and RoboArena~\cite{atreya2025roboarena}.
After filtering the data, we obtain $62{,}877$ DROID and $5{,}856$ RoboArena diverse tabletop manipulation trajectories.
DROID comprises $86$ diverse manipulation tasks across $564$ scenes and $52$ buildings at $13$ institutions, covering a broad range of everyday objects, tasks, and environments.
Unlike prior work that trains only on DROID~\cite{assran2025vjepa2, murlabadia2026vjepa21}, we include data from RoboArena, collected by rolling out different policies, to expose the model to a broader distribution of behaviors, scenes, and failure modes than DROID alone.
Each batch samples 8-frame video clips with synchronized side- and wrist-view observations at 4~FPS from $1280{\times}720$ images and resizes them to $256{\times}256$ for V-JEPA 2 encoders and $224{\times}224$ for DINOv3 encoders.
At each time step, the predictor is conditioned on a 7D end-effector action~$\mathbf{a}_t = \mathbf{s}_{t+1} - \mathbf{s}_t = [\Delta x, \Delta y, \Delta z, \Delta r, \Delta p, \Delta y, \Delta g]$,
where $\Delta x,\Delta y,\Delta z$ denote Cartesian translation, $\Delta r,\Delta p,\Delta y$ denote changes in end-effector roll, pitch, and yaw, and $\Delta g$ is the change in gripper state.

We train baseline world-model predictors that vary in visual encoder, camera view, and architecture. 
Specifically, we consider independently trained single-view predictors for the side and wrist cameras, alongside our proposed DUET dual-view predictor. 
Each predictor architecture is evaluated with two frozen pretrained visual encoder families: V-JEPA~2~\cite{assran2025vjepa2}, using a ViT-G/16 backbone with $1$B parameters, and DINOv3~\cite{simeoni2025dinov3}, using a ViT-H+/16 backbone with $840$M parameters.
We optimize all world-model variants using AdamW for $120$k steps with a batch size of $256$ with a learning rate of $4.25\times10^{-4}$ and a weight decay of $0.04$. 
The learning rate is warmed up from $7.5\times10^{-5}$ over the first $4{.}5$k steps and annealed to zero over the final $30$k steps.
The total computational cost of DUET-DINO over $120$k training steps across four GPUs is $5\times10^{20}$ FLOPs.
The combined training cost of all baseline and ablation models, excluding DUET-DINO, is $2.8\times10^{21}$ FLOPs.


\subsection{Experimental Setup}
\label{sec:exp_setup}
We evaluate latent-world-model planning on manipulation tasks that require full 7-DoF end-effector control.
These tasks include \emph{reach}, \emph{angled reach}, \emph{angled grasp and angled lift-to-home}. 
We organize the baseline and proposed world models into three predictor categories:
\begin{enumerate}
    \item \textbf{Single-view:} These models predict latent representations for either the side view or the wrist view using a single view-specific predictor.

    \item \textbf{Independent dual-view:} This baseline combines separately trained side- and wrist-view predictors, without cross-attention or joint optimization between the two.

    \item \textbf{DUET dual-view:} Our proposed model jointly trains two view-specific predictor heads together with cross-view attention blocks.
\end{enumerate}
In addition, we evaluate all tasks using the official V-JEPA~2-AC$^\ast$ checkpoint~\cite{assran2025vjepa2}, which was trained exclusively on the left side-view DROID data for $94.5$k optimization steps.
Fig.~\ref{fig:latent_planning} shows an overview of the prediction and planning pipeline.
Goal observations from the corresponding camera views specify each task.
For single-view predictors, the goal is provided from either the static side camera or the wrist camera, whereas dual-view predictors use paired side- and wrist-view goal observations.
Single-view predictors plan using the corresponding normalized view-specific cost, while dual-view predictors use the combined planning cost, as described in Sec.~\ref{sec:worldmodel_planning}.
Unlike prior work~\cite{assran2025vjepa2, murlabadia2026vjepa21}, which uses 3-DoF end-effector actions with gripper control, our planner samples candidate action sequences over the full 7-DoF end-effector action space.
For all tasks in simulation, planning consistently uses $800$ action samples, weighted $50$ top-$k$ candidates, and $2$ CEM iterations.
For each evaluation run, we use a run-specific CEM seed shared across all model variants, while different runs use different seeds.

\label{sec:sim_exp}
The proposed tasks are evaluated in the RoboLab simulator~\cite{yang2026robolab}, which provides visually realistic and diverse backgrounds, object assets, configurable camera views, and the DROID robot setup, making it well-suited for rigorous evaluation of latent planning for manipulation.
The DROID setup consists of a Franka arm equipped with a Robotiq 2F-85 gripper,
a wrist-mounted camera, and a static side-view camera.
Note that the right side-view camera is used consistently for both training and evaluation.
The end-effector actions produced by the planner are transformed into joint commands using inverse kinematics.

\begin{figure}[t]
\centering
\includegraphics[width=0.8\linewidth]{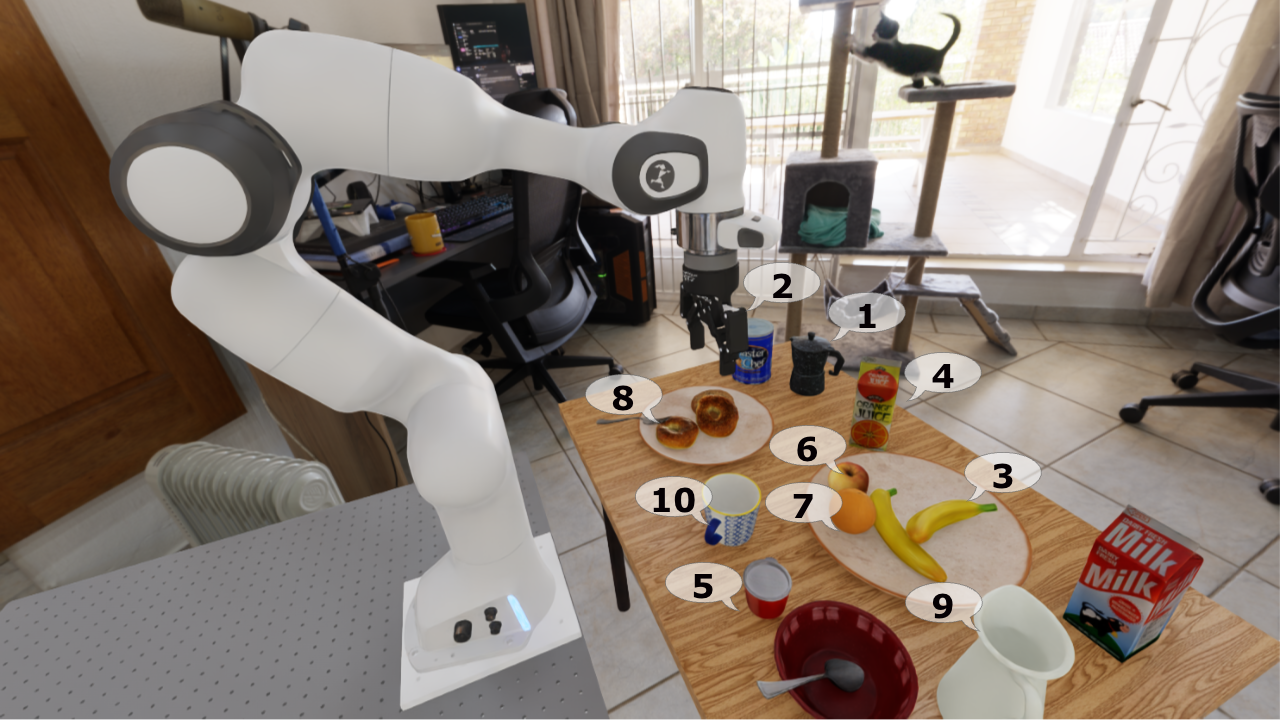}
\caption{Static right side-camera view showing the robot in its home pose and the target objects used in the reach tasks: (1) Coffee Pot, (2) Coffee Can, (3) Banana, (4) Juice Carton, (5) Yogurt Cup, (6) Apple, (7) Orange, (8) Bagel, (9) White Pitcher, and (10) Ceramic Mug.}
\label{fig:breakfast_table_reach_tasks}
\end{figure}
\subsection{Reach Task}
The experimental setup for the reach task consists of ten semantically distinct target objects distributed across the~\texttt{breakfast\_table} scene with a~\texttt{HomeOfficeBackground}, as shown in Fig.~\ref{fig:breakfast_table_reach_tasks}.
We define a reach run to be successful if the end-effector position~$p_\text{e}$ is within a distance of~$\epsilon_p$ to the goal position~$p_\mathrm{g}$, in $100$ steps, i.e., if~$\|p_\text{e} - p_\text{g}\|_2 \leq \epsilon_p$.
We define $\epsilon_p = 5\,\mathrm{cm}$ for all items.
The final position error (FPE) is the Euclidean distance computed at the final planning step, regardless of task success, and therefore includes residual error after successful task completion.

\begin{table}[t]
    \centering
    \setlength{\tabcolsep}{8pt}
    \renewcommand{\arraystretch}{1.}
    \caption{
    Aggregate reach success rate (SR) and final position error (FPE) over $100$ evaluation runs per model across ten tasks.
    }
    \label{tab:reach_aggregate}
    \begin{tabular}{cc l cc}
        \toprule
        \textbf{Side}
        & \textbf{Wrist}
        & \textbf{Predictor}
        & \makecell{\textbf{SR} \textbf{(\%)} \textbf{$\uparrow$}}
        & \makecell{\textbf{FPE} \textbf{(cm) $\downarrow$}} \\
        \midrule

        \rowcolor{gray!5}
        \multicolumn{5}{l}{\textbf{V-JEPA 2 Encoder}} \\

        $\checkmark$
        & --
        & V-JEPA 2-AC$^\ast$
        & $55\%$
        & $18.6$\std{23.2} \\
        
        $\checkmark$
        & --
        & Single-view
        & $20\%$
        & $34.4$\std{17.9} \\

        --
        & $\checkmark$
        & Single-view
        & $37\%$
        & $32.5$\std{26.6} \\

        $\checkmark$
        & $\checkmark$
        & Independent
        & $37\%$
        & $32.5$\std{24.4} \\

        $\checkmark$
        & $\checkmark$
        & DUET-VJEPA
        & $41\%$
        & $39.0$\std{38.6} \\

        \midrule

        \rowcolor{gray!5}
        \multicolumn{5}{l}{\textbf{DINOv3 Encoder}} \\

        $\checkmark$
        & --
        & Single-view
        & $18\%$
        & $28.0$\std{15.6} \\

        --
        & $\checkmark$
        & Single-view
        & $79\%$
        & $13.1$\std{20.3} \\

        $\checkmark$
        & $\checkmark$
        & Independent
        & $78\%$
        & $9.8$\std{13.8} \\

        \rowcolor{teal!20}
        $\mathbf{\checkmark}$
        & $\mathbf{\checkmark}$
        & \textbf{DUET-DINO}
        & $\mathbf{92\%}$
        & $\mathbf{5.4}$\std{\mathbf{7.4}} \\

        \bottomrule
    \end{tabular}
\end{table}
Table~\ref{tab:reach_aggregate} shows the reach performance of single-view and dual-view predictors in latent planning. 
V-JEPA~2-AC$^\ast$ with our visual planning achieves a higher success rate than all trained V-JEPA~2 baselines, suggesting that longer training on more diverse data degraded spatial planning.
The trained side-view predictors achieve low success rates and high FPE, suggesting limited ability to spatially plan end-effector actions across different regions of the table.
Successful runs were primarily observed for target objects located near the robot's home pose, such as objects (10) and (5).
The V-JEPA~2 wrist-view predictor shows a moderate improvement over its side-view. 
In contrast, the DINOv3 wrist-view predictor shows a larger performance gain.
Independent dual-view models provide negligible gains over the corresponding wrist-view predictors, highlighting the difficulty of naively combining views.
The DUET-VJEPA predictor improves over its corresponding independent dual-view model in terms of success rate. 
Despite this improvement, its FPE remains high because, in failed reach~\texttt{(3)Banana} runs, the planned end-effector actions drive the robot substantially away from the target object, resulting in a large FPE.
Our DUET-DINO predictor achieves the strongest overall spatial planning performance, reaching a success rate of $92\%$. 
Notably, DUET-DINO also succeeds on the challenging reach~\texttt{(1)Coffee Pot} runs near the table corner, which requires coordinated translational and rotational motion under kinematic constraints, while the target is outside the wrist camera's field of view at the robot's home pose.
~\cite{assran2025vjepa2} reports $16\,\mathrm{s}$ per planning step for a single-view predictor using more CEM iterations, while DUET-DINO requires $15$--$17\,\mathrm{s}$ with two predictors in our experiment setup. 
Overall, dual-view models are roughly $2\times$ slower than single-view models, and V-JEPA~2 dual-view variants are about $1.3\times$ slower than the corresponding DINOv3 models.

\subsection{Angled Reach Task}
To evaluate orientation planning, we define a set of orientation-intensive angled-reach tasks.
We define an angled reach run as successful if the translational and rotational errors satisfy~$\|p_\text{e} - p_\text{g}\|_2 \leq \epsilon_p$ and~$
\theta(q_\text{e}, q_\text{g}) \leq \epsilon_\theta$,
where~$q_\text{e}$ and~$q_\text{g}$ are the actual and goal orientation expressed as quaternions, and the orientation error is given by~$\theta(q_\text{e}, q_\text{g})=
2\cos^{-1}\left(\left| q_\text{e} \cdot q_\text{g} \right|\right)$.
We set the tolerances to~$\epsilon_p = 0.05\,\mathrm{m}$ and $\epsilon_\theta = 0.21$ (corresponding to $12^\circ$), which must be satisfied within $100$ steps for a run to be counted as successful.
The final angular error (FAE) is computed at the final planning step, regardless of whether a run is successful.

\textit{Fixed Background:}
First, we define four orientation-intensive angled-reach tasks across four distinct tabletop scenes in a \texttt{HomeOfficeBackground}. 
Two tasks primarily require vertical reaching along the $z$-axis, followed by clockwise and counterclockwise yaw rotations, respectively.
The remaining two tasks require coordinated translational motion in ($x$, $y$, $z$), together with clockwise and counterclockwise yaw rotations.
\begin{table}[t]
    \centering
    \setlength{\tabcolsep}{4pt}
    \renewcommand{\arraystretch}{1.}
    \caption{
    Aggregate angled-reach success rate (SR), final position error (FPE), and final angular error (FAE) over $40$ evaluation runs per model across four tasks.
    }
    \label{tab:angled_reach_aggregate}
    \begin{tabular}{cc l ccc}
        \toprule
        \textbf{Side}
        & \textbf{Wrist}
        & \textbf{Predictor}
        & \makecell{\textbf{SR} \textbf{(\%) $\uparrow$}}
        & \makecell{\textbf{FPE} \textbf{(cm) $\downarrow$}}
        & \makecell{\textbf{FAE} \textbf{($^\circ$) $\downarrow$}} \\
        \midrule

        \rowcolor{gray!5}
        \multicolumn{6}{l}{\textbf{V-JEPA 2 Encoder}} \\

        $\checkmark$
        & --
        & V-JEPA 2-AC$^\ast$
        & $2.5\%$
        & $3.6$\std{1.5} 
        & $72.6$\std{32.2}\\
        
        $\checkmark$
        & --
        & Single-view
        & $25.0\%$
        & $3.6$\std{1.6}
        & $51.9$\std{31.9} \\

        --
        & $\checkmark$
        & Single-view
        & $25.0\%$
        & $36.7$\std{22.5}
        & $67.9$\std{37.7} \\

        $\checkmark$
        & $\checkmark$
        & Independent
        & $25.0\%$
        & $18.4$\std{15.7}
        & $66.7$\std{36.8} \\

        $\checkmark$
        & $\checkmark$
        & DUET-VJEPA
        & $25.0\%$
        & $15.7$\std{20.6}
        & $69.7$\std{41.7} \\

        \midrule

        \rowcolor{gray!5}
        \multicolumn{6}{l}{\textbf{DINOv3 Encoder}} \\

        $\checkmark$
        & --
        & Single-view
        & $25.0\%$
        & $5.3$\std{1.5}
        & $67.1$\std{40.2} \\

        --
        & $\checkmark$
        & Single-view
        & $62.5\%$
        & $8.3$\std{7.2}
        & $47.4$\std{57.5} \\

        $\checkmark$
        & $\checkmark$
        & Independent
        & $55.0\%$
        & $3.2$\std{2.5}
        & $48.8$\std{54.4} \\

        \rowcolor{teal!20}
        $\mathbf{\checkmark}$
        & $\mathbf{\checkmark}$
        & \textbf{DUET-DINO}
        & $\mathbf{72.5\%}$
        & $\mathbf{1.9}$\std{\mathbf{1.2}}
        & $\mathbf{22.7}$\std{\mathbf{32.2}} \\

        \bottomrule
    \end{tabular}
\end{table} 
As shown in Table~\ref{tab:angled_reach_aggregate}, V-JEPA~2-AC$^\ast$ achieves a lower success rate than the trained V-JEPA~2 side-view baseline on angled-reach tasks, suggesting that while prior work performs well for coarse translational planning, it struggles with fine-grained control.
The trained side-view predictors fail on the most rotation-intensive angled-reach tasks and succeed only on the simpler rotation cases.
Since the V-JEPA~2 wrist-view predictor also performs poorly, the corresponding independent dual-view and DUET-VJEPA predictors show no improvement.
The DINOv3 wrist-view predictor shows substantially stronger performance, suggesting that DINOv3 latent features capture orientation changes effectively and that the corresponding action-conditioned predictor learns these dynamics well.
Although the DINOv3 independent dual-view model improves FPE by leveraging the side-view predictor, its FAE increases when both views are combined, resulting in a lower success rate.
\algname{} reduces both final position and angular errors, resulting in a higher success rate.

\textit{Diverse Backgrounds and Distractor Objects:}
\begin{table}[tb!]
    \centering
    \setlength{\tabcolsep}{4.4pt}
    \renewcommand{\arraystretch}{1.}
    \caption{
        Aggregate angled-reach success rate (SR), final position error (FPE), and final angular error (FAE) under visual distribution shifts over $100$ evaluation runs per model.
    }
    \label{tab:angled_reach_bgdistractors}

    \begin{tabular}{cc l ccc}
        \toprule
        \textbf{Side}
        & \textbf{Wrist}
        & \textbf{Predictor}
        & \makecell{\textbf{SR} \textbf{(\%) $\uparrow$}}
        & \makecell{\textbf{FPE} \textbf{(cm) $\downarrow$}}
        & \makecell{\textbf{FAE} \textbf{($^\circ$) $\downarrow$}} \\
        \midrule

        \rowcolor{gray!5}
        \multicolumn{6}{l}{\textbf{DINOv3 Encoder}} \\

        --
        & $\checkmark$
        & Single-view
        & $25\%$
        & $15.0$\std{10.0}
        & $54.3$\std{42.4} \\

        $\checkmark$
        & $\checkmark$
        & Independent
        & $45\%$
        & $3.3$\std{2.4}
        & $47.3$\std{40.8} \\

        \rowcolor{teal!20}
        $\mathbf{\checkmark}$
        & $\mathbf{\checkmark}$
        & \textbf{DUET-DINO}
        & $\mathbf{63\%}$
        & $\mathbf{2.6}$\std{\mathbf{2.1}}
        & $\mathbf{30.9}$\std{\mathbf{36.5}} \\

        \bottomrule
    \end{tabular}
\end{table}
Next, we rigorously evaluate the best-performing models on the same orientation-intensive tasks under visual distribution shifts across five diverse backgrounds and tabletop objects. 
For each of five backgrounds, we vary the tabletop scene from one target object to five tabletop objects, including visually similar distractors. 
We evaluate these variations across four angled-reach tasks, resulting in \(5 \times 5 \times 4 = 100\) evaluation runs per model.
Table~\ref{tab:angled_reach_bgdistractors} shows that \algname{} remains robust under these visual variations, while the wrist-view predictor exhibits a noticeable drop in success rate.
We observe that the overall angled-reach planning performance is particularly strong for asymmetrically shaped target objects.

\textit{Hardware Runs:}
\begin{table}[t]
    \centering
    \setlength{\tabcolsep}{4pt}
    \renewcommand{\arraystretch}{1.}
    \caption{
Aggregate angled-reach success rates (SR), final position errors (FPE), and final angular errors (FAE) over $30$ hardware evaluation runs across three tasks.
    }
    \label{tab:angled_reach_hw}

    \begin{tabular}{cc l ccc}
        \toprule
        \textbf{Side}
        & \textbf{Wrist}
        & \textbf{Predictor}
        & \makecell{\textbf{SR} \textbf{(\%) $\uparrow$}}
        & \makecell{\textbf{FPE} \textbf{(cm) $\downarrow$}}
        & \makecell{\textbf{FAE} \textbf{($^\circ$) $\downarrow$}} \\
        \midrule

        \rowcolor{gray!5}
        \multicolumn{6}{l}{\textbf{DINOv3 Encoder}} \\

        --
        & $\checkmark$
        & Single-view
        & $3.3\%$
        & $24.9$\std{21.0}
        & $39.9$\std{31.5} \\

        $\checkmark$
        & $\checkmark$
        & Independent
        & $16.7\%$
        & $\mathbf{14.3}$\std{\mathbf{17.2}}
        & $40.0$\std{42.0} \\

        \rowcolor{teal!20}
        $\mathbf{\checkmark}$
        & $\mathbf{\checkmark}$
        & \textbf{DUET-DINO}
        & $\mathbf{26.7\%}$
        & $14.7$\std{17.1}
        & $\mathbf{28.0}$\std{\mathbf{17.7}} \\

        \bottomrule
    \end{tabular}
\end{table}
Finally, we evaluate the best-performing DINOv3-based predictors on three angled-reach tasks using a real DROID robot setup. 
We selected challenging target poses near the table corners with both clockwise and counterclockwise target orientations.
Our DROID hardware setup consists of a table-mounted Franka Research 3 robot equipped with a Robotiq 2F-85 gripper, a ZED 2i camera for the static side view, and a ZED Mini camera for the wrist view.
We use the Robot Control Stack~\cite{julg2025robot} for real-world robot control and VLAgents~\cite{juelg2026vlagentspolicyserverefficient} to serve the policy.
Results in Table~\ref{tab:angled_reach_hw} show that DUET-DINO achieves the highest success rate on the challenging angled-reach poses, outperforming the independent dual-view predictor.
Due to the computational cost and latency of CEM planning on hardware, we use one CEM iteration with $500$ candidate actions and $15$ planning steps.
This reduced planning budget contributes to the lower hardware success rates relative to the simulation results.
For safety, actions resulting in collisions or kinematic singularities were rejected, and runs were terminated if unsafe actions were proposed for three consecutive planning steps. 
The wrist-view predictor produced unsafe actions more frequently, further contributing to its lower success rate on hardware.

\subsection{Angled Grasp and Angled Lift-to-Home}
We further extend the angled-reach tasks in the \texttt{HomeOfficeBackground} with grasping and lifting subtasks, thereby evaluating the models on sequential multi-goal planning.
We allocate $60$ planning steps for angled reaching (successful if the pose error is below $15\,\mathrm{cm}$ and $20^\circ$), $10$ steps for grasping, and $60$ steps for lifting the grasped object to the initial robot home pose (successful if the end-effector position is within $20\,\mathrm{cm}$ of the home position while maintaining the grasp).
Most of the \algname{} runs succeed in all sub-tasks, thereby completing the full lift task. 
The DINOv3 wrist-view predictor often succeeds at angled reaching and grasping but loses task context during the lift-to-home phase, resulting in lift failure.
\begin{table}[t]
    \centering
    \setlength{\tabcolsep}{8pt}
    \renewcommand{\arraystretch}{1.}
    \caption{
    Aggregate stage-wise success rates (SR) for angled-lift tasks over $40$ evaluation runs across four tasks.
    }
    \label{tab:angled_pick}
    \begin{tabular}{cc l cc}
        \toprule
        \textbf{Side}
        & \textbf{Wrist}
        & \textbf{Predictor}
        & \makecell{\textbf{Angled-Reach} \\ \textbf{SR (\%) $\uparrow$}}
        & \makecell{\textbf{Lift} \\ \textbf{SR} \textbf{(\%) $\uparrow$}} \\
        \midrule

        \rowcolor{gray!5}
        \multicolumn{5}{l}{\textbf{DINOv3 Encoder}} \\

        --
        & $\checkmark$
        & Single-view
        & $57.5\%$
        & $20.0\%$ \\

        $\checkmark$
        & $\checkmark$
        & Independent
        & $60.0\%$
        & $57.5\%$ \\

        \rowcolor{teal!20}
        $\mathbf{\checkmark}$
        & $\mathbf{\checkmark}$
        & \textbf{DUET-DINO}
        & $\mathbf{75.0\%}$
        & $\mathbf{60.0\%}$ \\

        \bottomrule
    \end{tabular}
\end{table}

\subsection{Ablation}
\textit{Cross-view Cross Attention:}
We train an ablated \algname{} variant without the cross-attention blocks.
This reduces the reach performance of \algname{} in Table~\ref{tab:reach_aggregate} to $81\%$ SR with $10.7$\std{15.4}$\,\mathrm{cm}$ FPE, and the angled-reach performance in Table~\ref{tab:angled_reach_aggregate} to $42.5\%$ SR with $4.3$\std{3.1}$\,\mathrm{cm}$ FPE and $55.5$\std{52.8}$^\circ$ FAE. This confirms the importance of cross-view attention for fine-grained dual-view planning.

\textit{Camera Perturbations:}
The side-view predictor learns action-conditioned visual dynamics from a fixed camera viewpoint, i.e., robot joint motion during reaching and object motion in contact-rich interactions. 
We therefore evaluate robustness to camera-pose shifts on \texttt{ReachBananaTask} over $30$ runs, perturbing the camera by $\pm20\,\mathrm{cm}$ in $x/y$, $\pm10\,\mathrm{cm}$ in $z$, and $\pm11.5^\circ$ in roll, pitch, and yaw, using matched perturbation seeds across models' runs.
Under these shifts, DUET-DINO and the DINOv3 independent dual-view predictor achieve $66.7\%$ SR, compared with $46.7\%$ for DUET-VJEPA and $40.0\%$ for the V-JEPA~2 independent dual-view predictor, while both single-view side predictors fail with $0\%$ SR.
These results highlight the sensitivity of side-view planning to camera-pose shifts and the robustness gained from incorporating the wrist view.

\textit{Training Data:}
To study the effect of training data, we train DUET-DINO on DROID only for $94.5$k steps following the prior work~\cite{assran2025vjepa2}. 
The DROID-only DUET-DINO achieves $78\%$ SR on the reach tasks in Table~\ref{tab:reach_aggregate}, showing the benefit of additional training data on spatial planning. 
On the angled-reach tasks in Table~\ref{tab:angled_reach_aggregate}, the DROID-only DUET-DINO achieves $72.5\%$ SR, showing no improvement in orientation planning.
 
\begin{figure}
\centering
\includegraphics[width=\linewidth]{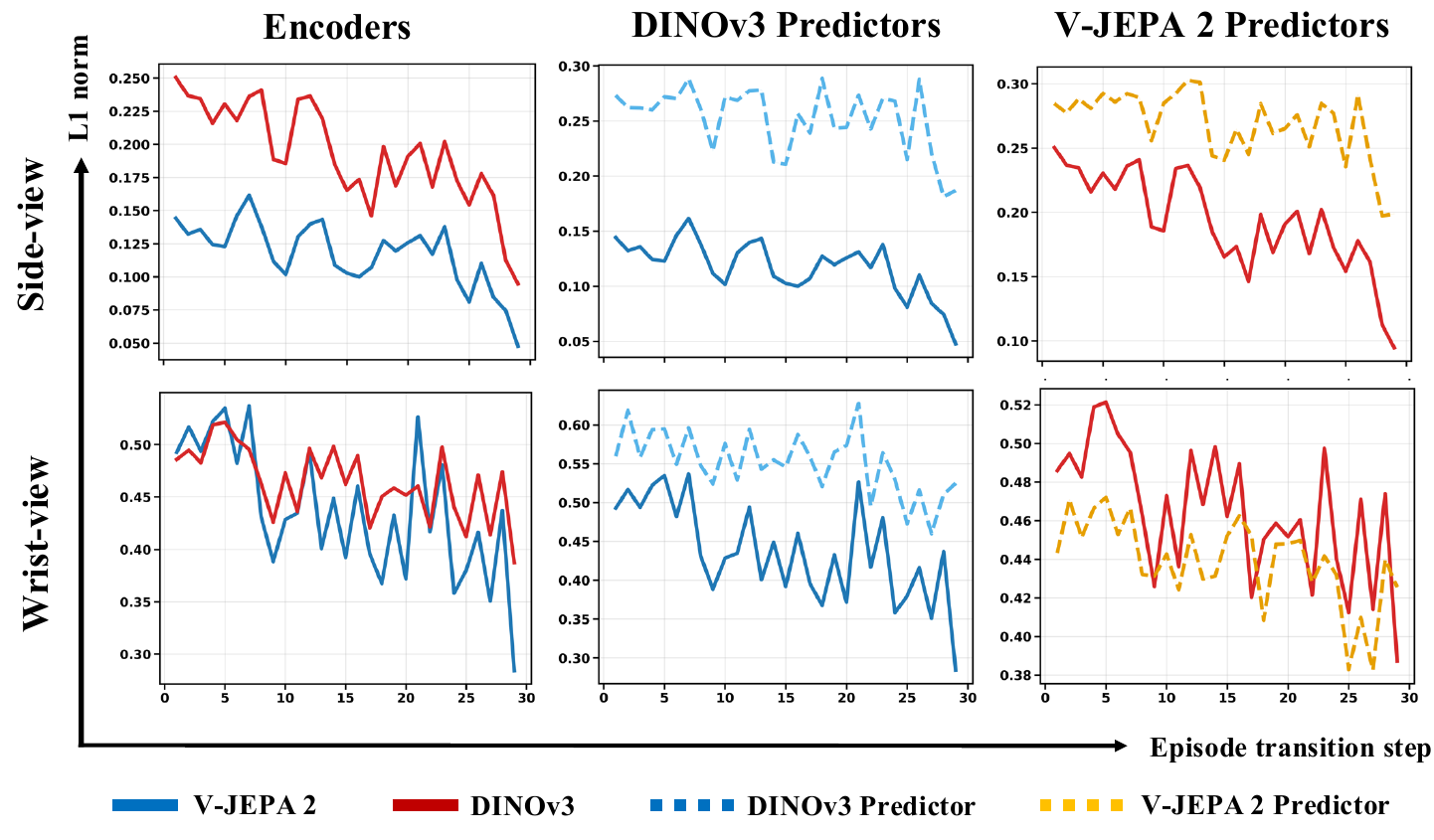}
\caption{Mean latent changes captured by frozen encoders, DINOv3 action-conditioned predictors, and V-JEPA~2 action-conditioned predictors over 30 runs of an angled-reach task.}
\label{fig:latent_analysis}
\end{figure}
\begin{figure}[t]
\centering
\includegraphics[width=\linewidth]{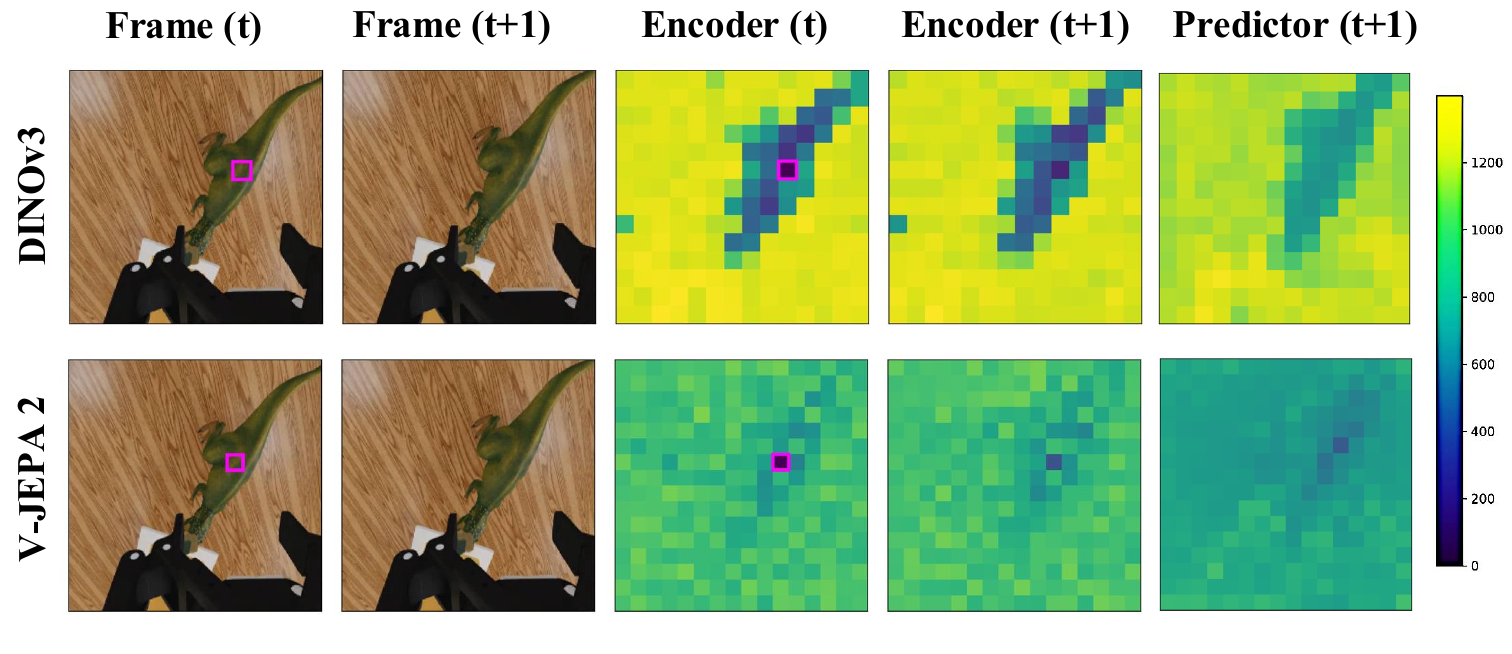}
\caption{
Patch-wise latent distances for DINOv3 and V-JEPA~2 feature maps.
Lower $\ell_1$ distances indicate greater similarity to the selected reference patch.
Frames at $t$ and $t+1$ are shown after resizing to the respective encoder inputs.  
Feature maps are visualized at $14\times14$ for DINOv3 and $16\times16$ for V-JEPA~2. 
DINOv3 produces $196$ patches of $1280$ dimensions, whereas V-JEPA~2 produces $256$ patches of $1408$ dimensions.
}
\label{fig:PatchWise_l1_norm}
\end{figure}
\subsection{Embeddings Analysis}
To understand why DINOv3-based predictors outperform those based on V-JEPA~2, we analyze the underlying latent representations.
Using an angled-reach rollout, we encode all frames with the frozen DINOv3 and V-JEPA~2 encoders and compute the $\ell_1$ norm between consecutive feature maps.
This measures the sensitivity of each representation space to visual changes over action steps.
As shown in Fig.~\ref{fig:latent_analysis}, first column, V-JEPA~2 and DINOv3 encoder features exhibit latent changes of similar magnitude, with slightly larger side-view changes for V-JEPA~2.
We then measure the scene dynamics produced by the predictors by computing the $\ell_1$ distance between the encoded input features and the action-conditioned predicted next-step features.
In the second column, the DINOv3 side- and wrist-view predictions follow the ground-truth temporal pattern, while generally overestimating their magnitude due to prediction error.
In contrast, the last column shows that the V-JEPA~2 wrist-view predictor underestimates latent changes, while the side-view predictor better captures the magnitude, suggesting weaker modeling of fine-grained action-conditioned dynamics by V-JEPA~2 wrist predictor.

In Fig.~\ref{fig:PatchWise_l1_norm}, we analyze wrist-view patch correspondences by selecting a target-object latent patch at time $t$ and computing its $\ell_1$ distance to all patches at $t$, $t+1$, and the predicted $t+1$ feature maps.
Lower distances indicate higher feature similarity.
DINOv3 embeddings show coherent correspondences over semantically related object regions, and its AC-predictor closely preserves the encoded $t+1$ pattern.
In contrast, V-JEPA~2 embedding exhibits noisier correspondences, with predicted low-distance regions also appearing in semantically unrelated areas.


%% file: 6_conclusion.tex
\section{Conclusion}
\label{sec:conclusion}
We present DUET-DINO, a cross-view latent world model for action-conditioned prediction and planning in robot manipulation.
By combining cross-view latent conditioning with DINO representations, DUET-DINO provides an effective approach for modeling scene changes induced by translational, rotational, and gripper actions. 
The evaluations are computationally expensive as CEM evaluates many candidate actions through world models at each step. 
While this enables rigorous world-model evaluation under diverse action proposals, it limits real-time control. 
Our future work will leverage DUET-DINO on a broader range of real-world tasks and use action proposals from generalist robot policies, such as vision-language-action (VLA) models, to reduce the planning search space and enable faster closed-loop control.

\section{Acknowledgements}
We thank Volker Schneider for support with the hardware setup and 3D-printing, and Prasanna Bhat for helping with real-world evaluations.
This work has been partially supported by the German Federal Ministry of Research, Technology and Space (BMFTR) under the Robotics Institute Germany (RIG).
Ralf Römer gratefully acknowledges support from the German Research Foundation (DFG) within the RTG project ConVeY, funded by grant GRK 2428.
The authors acknowledge the HPC resources
provided by the Erlangen National HPC Center (NHR@FAU)
under the BayernKI project no. v106be.